\documentclass[10pt,twocolumn,letterpaper]{article}
\pdfoutput=1
\usepackage{cvpr}
\usepackage{caption}
\usepackage{url}
\usepackage{cite}
\usepackage{amsmath}
\usepackage{amssymb}
\usepackage{times}
\usepackage{epsfig}
\usepackage{graphicx}
\usepackage{amsmath}
\usepackage{amssymb}
\usepackage{tabularx}
\usepackage{multirow}
\usepackage{subfigure}
\usepackage{comment}
\usepackage{float}
\usepackage{wrapfig}
\usepackage[ruled,vlined]{algorithm2e}
\usepackage{algorithmic}
\usepackage{bm}
\usepackage{tabu}
\usepackage{xspace}
\usepackage{booktabs}
\makeatletter
\DeclareRobustCommand\onedot{\futurelet\@let@token\@onedot}
\def\@onedot{\ifx\@let@token.\else.\null\fi\xspace}

\def\etal{et al\onedot}
\usepackage{xcolor}

\newcommand{\fig}[1]{figure~\ref{fig:#1}}

\usepackage{booktabs}

\usepackage[pagebackref=true,breaklinks=true,letterpaper=true,colorlinks,bookmarks=false]{hyperref}
\cvprfinalcopy 
\def\cvprPaperID{162} 

\ifcvprfinal\fi
\begin{document}

\title{Video Acceleration Magnification}
\author{
Yichao Zhang,  Silvia L. Pintea, and Jan C. van Gemert\\
Vision Lab, Delft University of Technology\\
Delft, Netherlands\\
}
\twocolumn[{%
\renewcommand\twocolumn[1][]{#1}%
\maketitle
\begin{center}
    \centering
	\begin{tabular}{cccc}
		\includegraphics[width=0.22\textwidth]{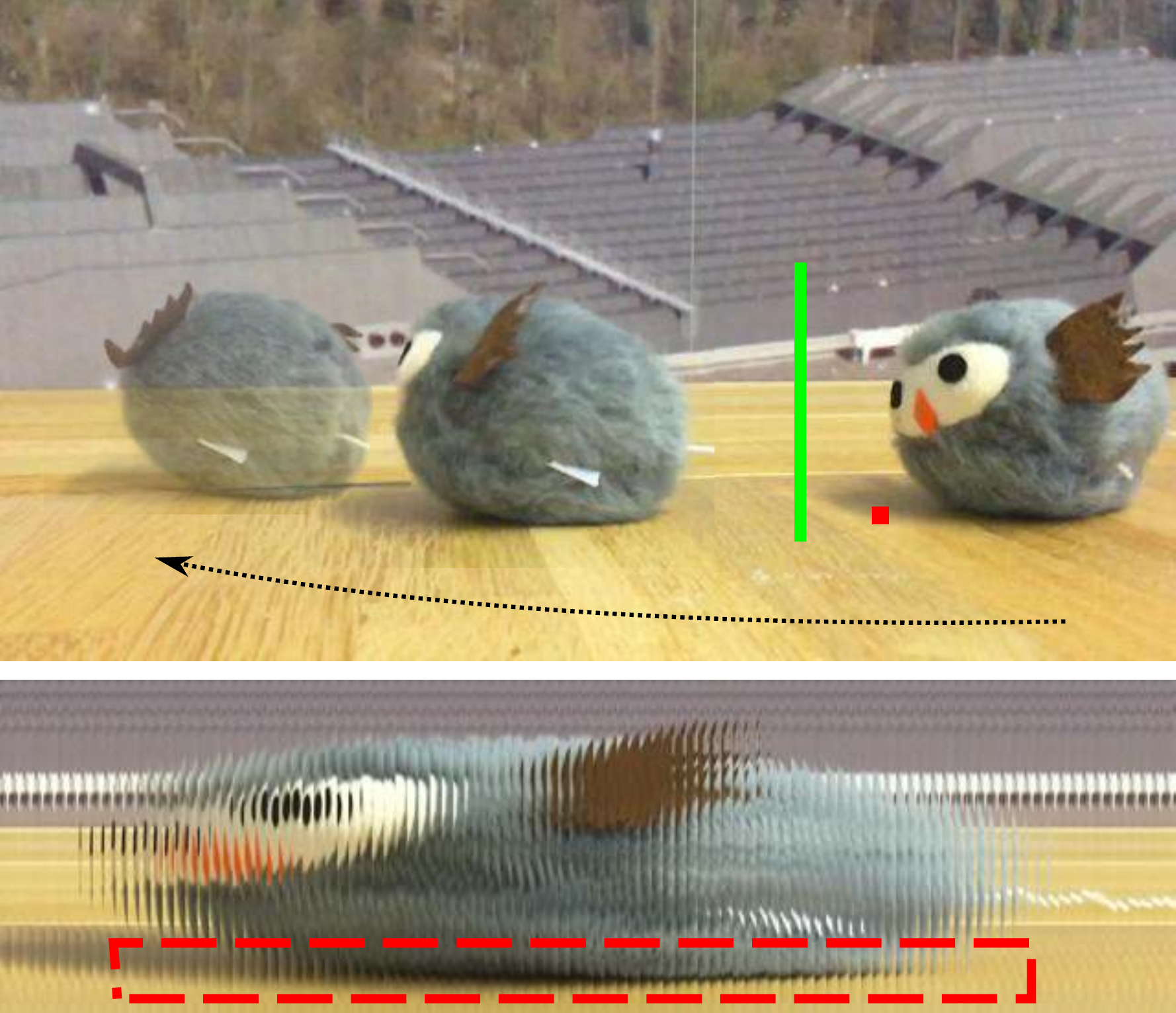} &
		\includegraphics[width=0.22\textwidth]{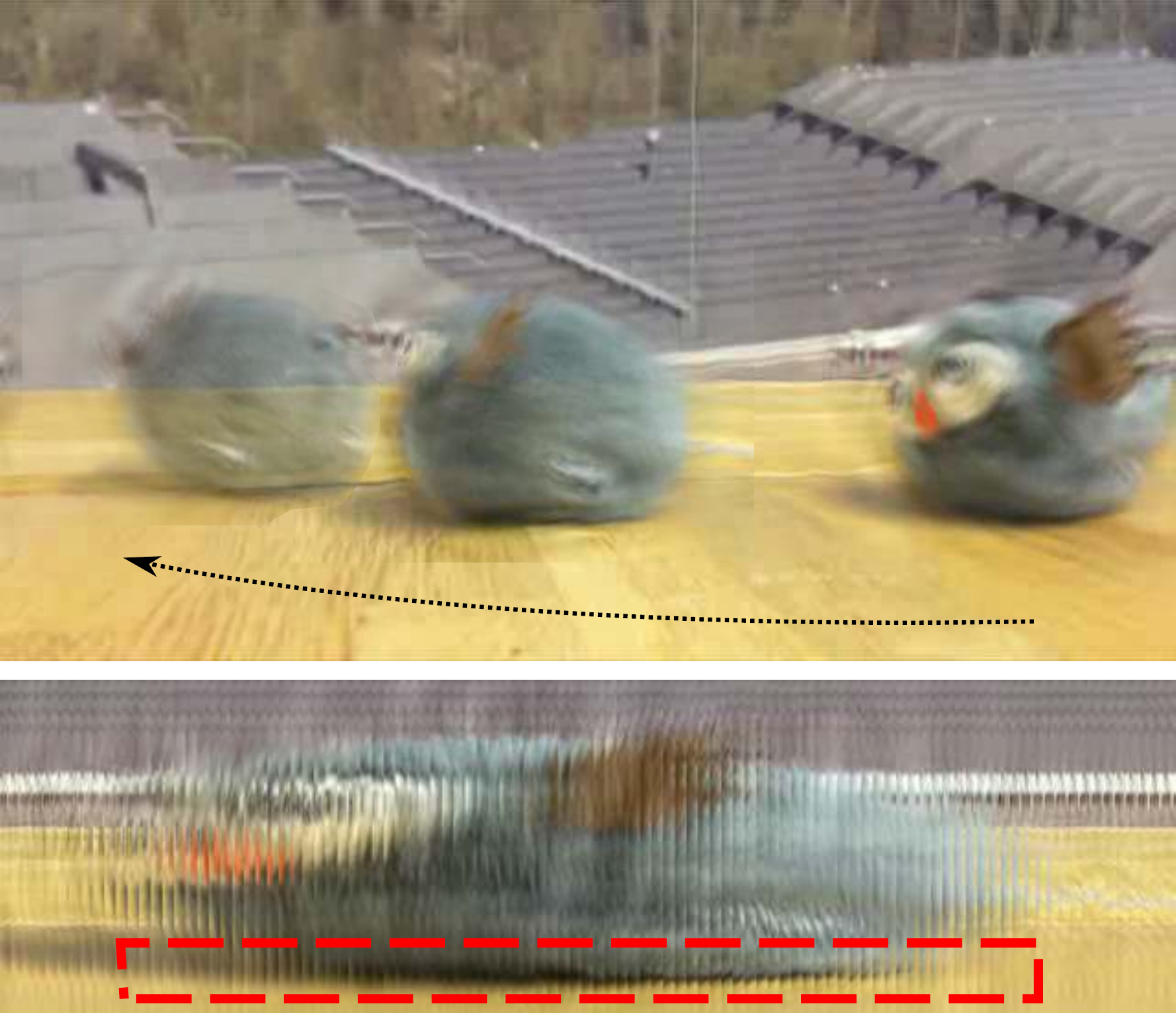} & 
		\includegraphics[width=0.22\textwidth]{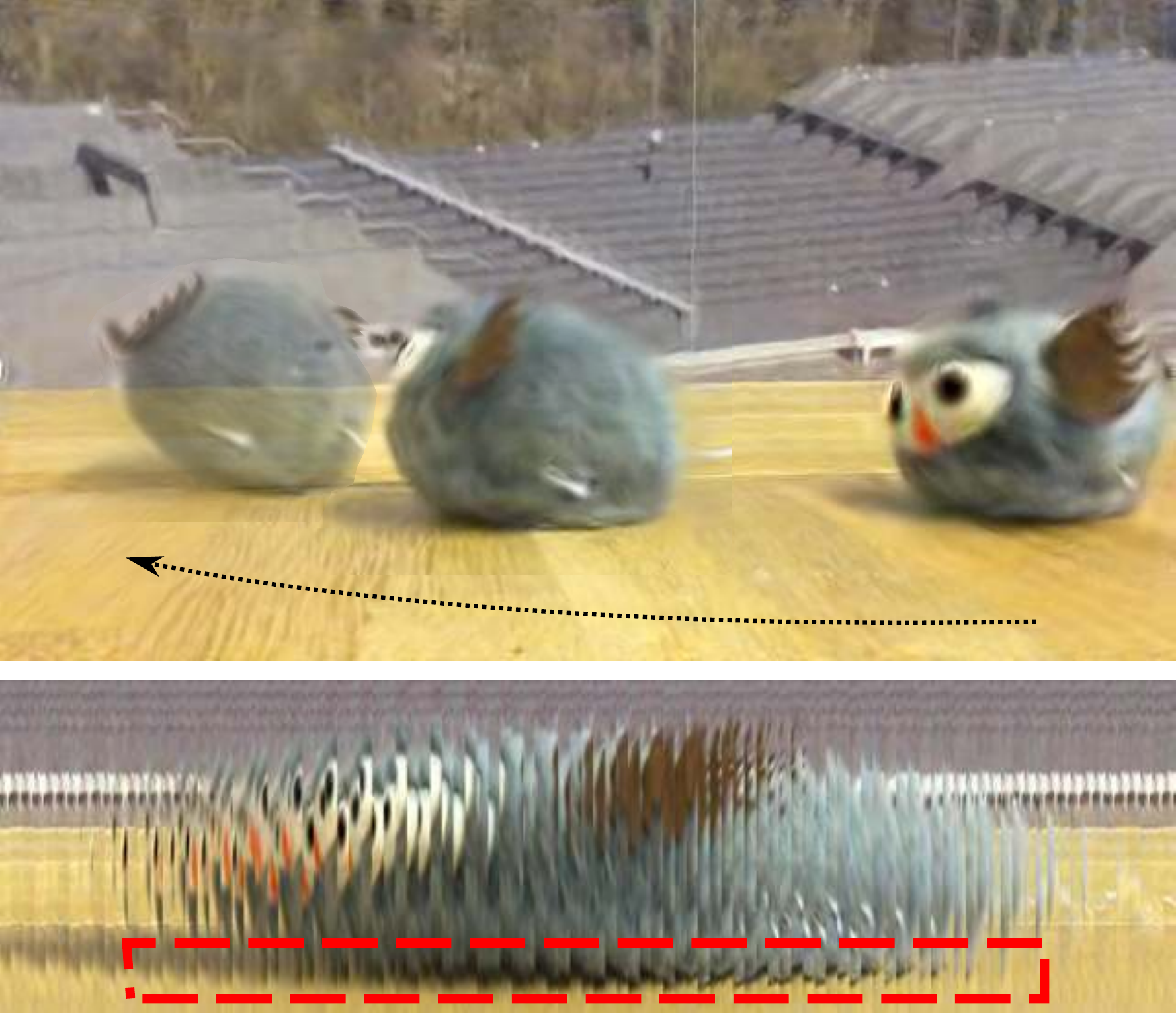} &
		\includegraphics[width=0.22\textwidth]{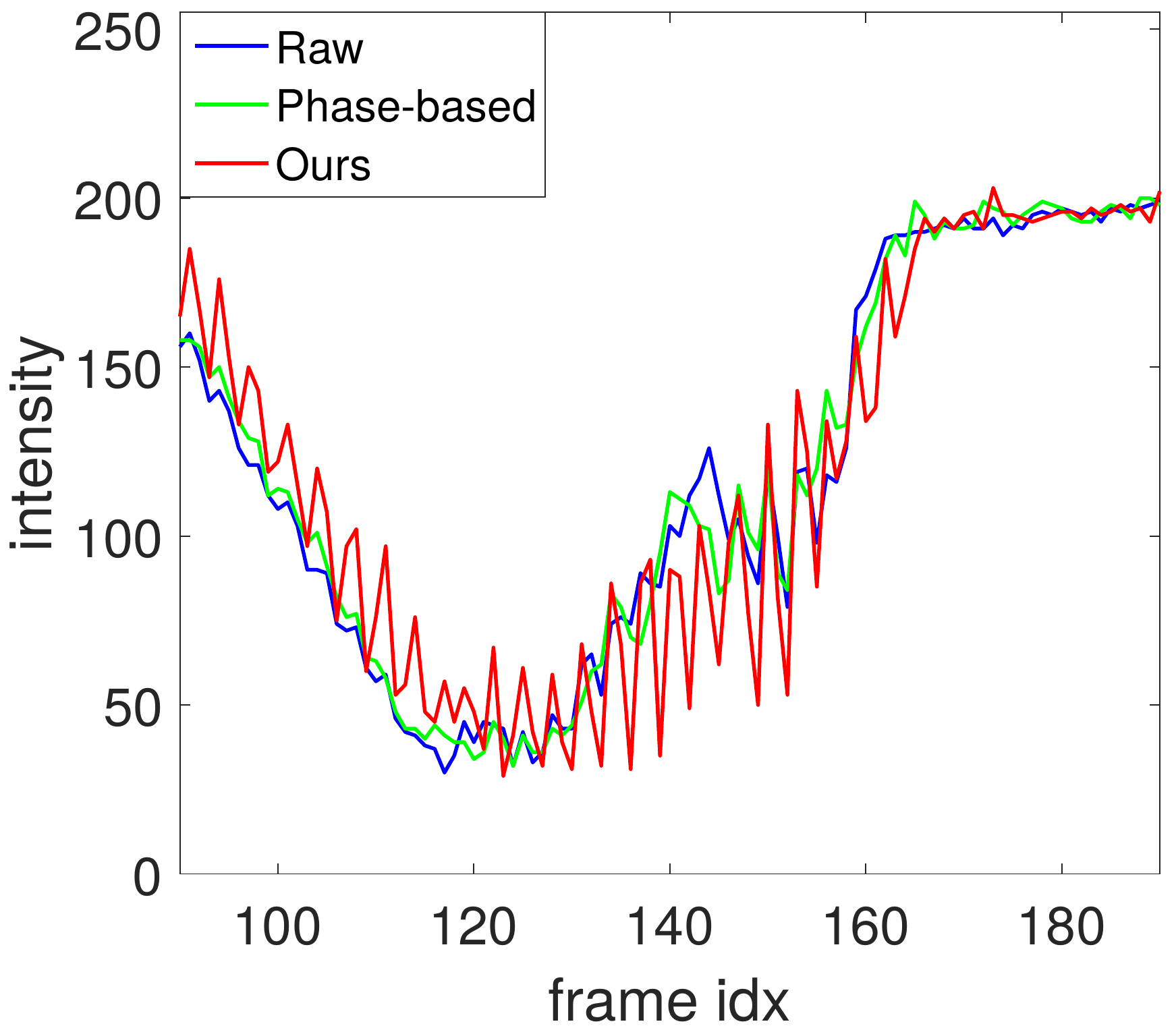} \\
		(a) Original video. & (b) Phase-based \cite{wadhwaSIGGRAPH13phase}. & (c) Ours. & (d) Intensity changes.\\
	\end{tabular}
    \captionof{figure}{\small 
	A toy moving along a trajectory depicted by the black arrow, while vibrating at a high frequency. 
	The top row shows 3 frames overlayed to indicate the toy's trajectory. 
	The bottom row shows a single column of pixels -- the green line in (a) -- for relevant video frames.  
	(a) Original video. 
	(b) Phase-based motion magnification \cite{wadhwaSIGGRAPH13phase}. 
	(c) Our proposed acceleration magnification.
	(d) Intensity changes at the location of the red pixel in the top row in (a) --- corresponding to a spatio-temporal rectangle in the bottom row.
	Our method generates sharper results with a greater magnification than the phase-based method in \cite{wadhwaSIGGRAPH13phase}. 
	See the supplementary material for the video result.}
	\label{fig:cattoy}
\end{center}%
}]
\thispagestyle{empty}


\begin{abstract}
The ability to amplify or reduce subtle image changes over time is useful in contexts such as video editing, medical video analysis, product quality control and sports. In these contexts there is often large motion present which severely distorts current video amplification methods that magnify change linearly. In this work we propose a method to cope with large motions while still magnifying small changes. We make the following two observations: i) large motions are linear on the temporal scale of the small changes; ii) small changes deviate from this linearity. We ignore linear motion and propose to magnify acceleration. Our method is pure Eulerian and does not require any optical flow, temporal alignment or region annotations. We link temporal second-order derivative filtering to spatial acceleration magnification. We apply our method to moving objects where we show motion magnification and color magnification. We provide quantitative as well as qualitative evidence for our method while comparing to the state-of-the-art.    
\end{abstract}

\vspace{-5mm}
\section{Introduction}

Essential properties of dynamic objects become clear only when they move. Consider, for example, the mechanical stability of a drone in flight, the muscles of an athlete doing sports, or the tremors of a Parkinson patient during walking. For these examples the properties of interest do not emerge while remaining still. The essential properties are the tiny variations that occur only during motion.

Tiny temporal variations that are hard or impossible to see with the naked eye can be enhanced by impressive video magnification algorithms~\cite{wuSIGGRAPH12eulerian,wadhwaSIGGRAPH13phase}. The strength of these methods stems from using Eulerian motion analysis instead of Lagrangian motion. The Lagrangian approach uses optical flow which is expensive and an unsolved research topic in its own~\cite{kroeger2016fast,fischer2015flownet,revaud2015epicflow}. Instead, the Eulerian approach does not require tracking; it measures flux at a fixed position. The Eulerian motion magnification methods~\cite{wuSIGGRAPH12eulerian,wadhwaSIGGRAPH13phase} give excellent results for magnifying blood flow, a heart-beat, or tiny breathing when the object and camera remain still. Unfortunately, these methods fail for moving objects because large motions overwhelm the small temporal variations.

A useful video magnification method that deals with large motion is developed by Elgharib \etal~\cite{elgharibCVPR15largeMotionMagnify}. It offers a hybrid of Eulerian and Lagrangian methods. By manually selecting the regions to magnify, these regions can be tracked by Lagrangian methods and subsequently temporally aligned using a homography. After alignment standard Eulerian magnification methods~\cite{wuSIGGRAPH12eulerian,wadhwaSIGGRAPH13phase} can be applied, yielding good magnification results. A disadvantage of this method is that regions of interest require manual segmentation which is time consuming and error prone. Also, the Lagrangian region tracking is expensive and sensitive to occlusions and 3D rotations. Furthermore, the alignment assumes a homography, which is often inaccurate for a non-static camera and non-planar objects. There is some room for improvement.

In this paper we propose video acceleration magnification for amplifying small variations in the presence of large motion. Our method does not require manual region annotation nor tracking or region alignment as done in~\cite{elgharibCVPR15largeMotionMagnify}. Instead, our method is closer to the original Eulerian approach~\cite{wuSIGGRAPH12eulerian} in its elegant simplicity. We make the observation that at the scale of the small variations the large motion is typically linear. By only magnifying small deviations of linear motion we arrive at accelerations magnification. 

The contributions of this paper are as follows. 1) We propose a pure Eulerian method for magnifying small variations in the presence of large motion. 2) We show the relation between a second-order temporal derivative filter and spatial acceleration magnification. 3) We give practical insight and analyze the success and failure of our method.
4) We outperform relevant video magnification baselines both in observed output quality and in a quantitative evaluation.

\section{Related Work}
\subsection{Lagrangian Approaches}
For the task of motion magnification, successful work focused on Lagrangian approaches.
These methods consider the image changes that happen over time at a certain object location by matching image points or patches between video frames 
and estimating the motion based on optical flow.   
In the presence of large object motion or camera motion, robust image registration plays a main role for such methods.
In \cite{liu2005motion} features are extracted over the frame and these features are tracked and clustered into groups of points where the video changes are magnified.
The work in \cite{balakrishnan2013detecting} estimates the heart beat of people from subtle movements of the head. 
It does so by extracting features over the head region and tracking them.
In more recent work on heart-rate estimation \cite{tulyakovECCV16heartRateMatrixCompletion} the tracking and selection of features is achieved by matrix completion.
The work in \cite{bai2012selectively} employs user input to define regions of large motion at which video de-animation is performed by 
tracking the pixels and using graph-cut to consistently segment the motion. 
Dissimilar to these works, we propose an Eulerian approach that does not rely on image registration, can deal with object and 
camera motion, and still magnifies the small video changes. 
\subsection{Eulerian Approaches}
Rather than the Lagrangian paradigm based on tracking points over time to estimate the changes of certain objects, 
the Eulerian paradigm analyzes the image changes over time at fixed image locations. 
Eulerian methods towards magnifying subtle video changes were proposed by first decomposing the video frames spatially through band-pass filtering, 
and then temporally filtering the signal to find the information to be magnified \cite{rubinstein2013revealing,wuSIGGRAPH12eulerian}.
These works have shown impressive results especially in the context of color amplification and heart rate estimation. 
With the apprise of the complex-steerable pyramid \cite{freeman1991design,simoncelli1995steerable,portilla2000parametric}, the use of phase-based motion 
has been considered not only in the context of motion magnification but also for other motion-related applications.
Examples include phase-based video frame interpolation \cite{meyer2015phase} and video modification transfer \cite{meyer2016phase}.
In \cite{davis2014visual} phase information is used for extracting sound from high speed cameras, 
while in \cite{davis2015visual} the video phase information is employed for predicting object material and in  ~\cite{chen2015modal} phase aids in estimating measurements of structural vibrations.
In the context of motion magnification, the successful work in \cite{wadhwaSIGGRAPH13phase} proposes the use of phase estimated through 
complex steerable filters and then magnifies this phase information. 
A speedup is proposed in \cite{wadhwaICCP2014riesz} through the use of a Riesz pyramid as an approximation for the complex pyramid.  
In the supplementary material of \cite{wuSIGGRAPH12eulerian} (Table 3, row 9) it is shown that the computational speed of an  acceleration filter can be improved.
These works achieve impressive results for motion magnification, however the downside of these approaches is that the subtle motion to be magnified must be isolated  --- 
no large object motion or camera motion should be present.  
Inspired by these works, we use a pure Eulerian approach to magnifying subtle video motion and we extend these methods to deal with large object or camera motion. 

To deal with camera and object motion, in \cite{elgharibCVPR15largeMotionMagnify}, the user is asked to indicate a 
frame region whose pixels are tracked and their motion is magnified. 
The recent work in \cite{kooij2016depth} proposes an alternative to finding the pixels whose changes should be magnified, 
by using depth cameras and bilateral filters such that the motion magnification is applied on all pixels located at the same depth.
However this method is not tested on moving objects.
Dissimilar to these works, we aim to perform video enhancement without the use of additional information such as user input or depth information.

\section{Acceleration Magnification}
\subsection{Linear Video Magnification}

\begin{figure*} 
\centering
\begin{tabular}{cc}
\centering
\includegraphics[width=0.48\textwidth]{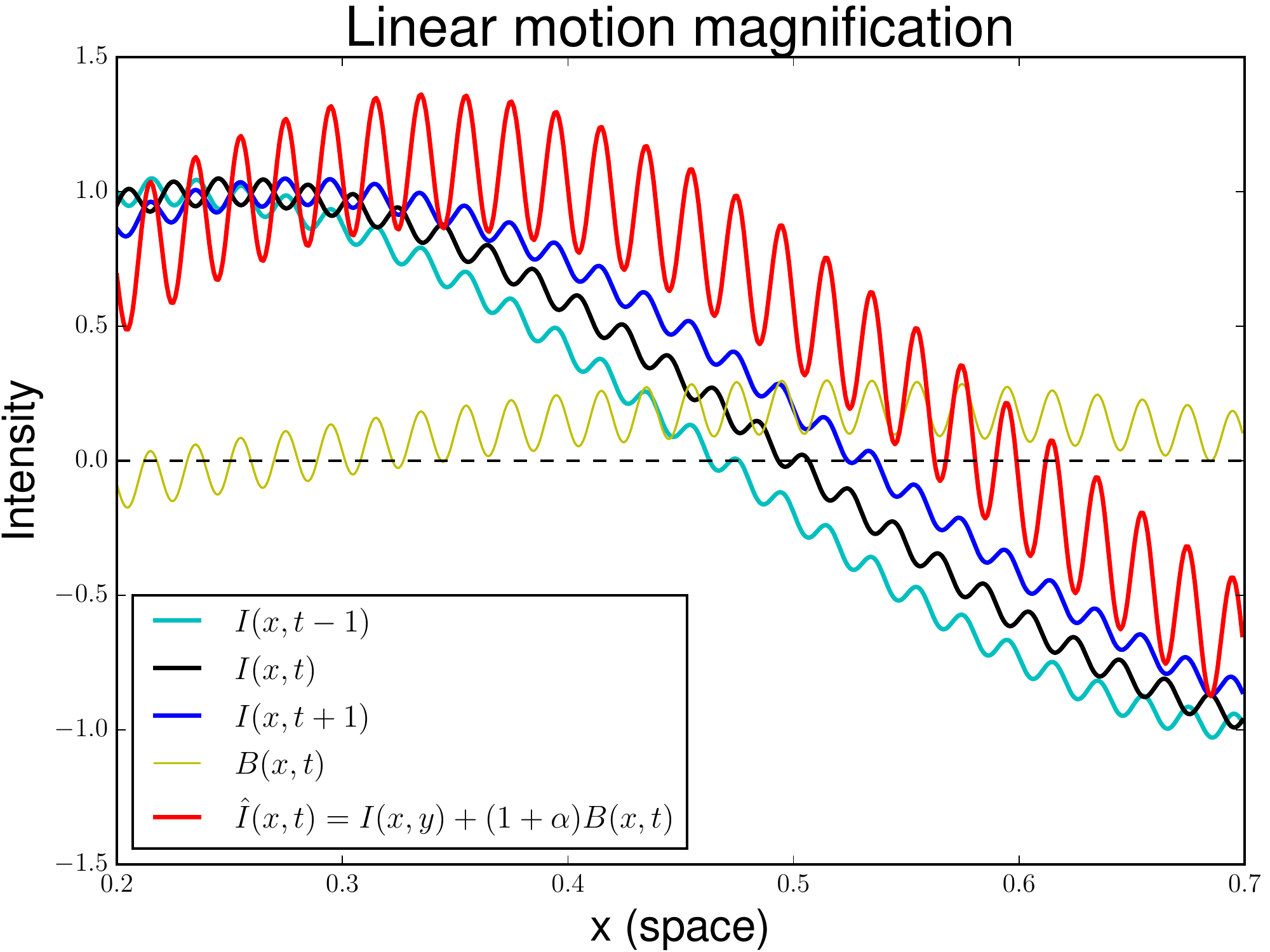} &
 \includegraphics[width=0.48\textwidth]{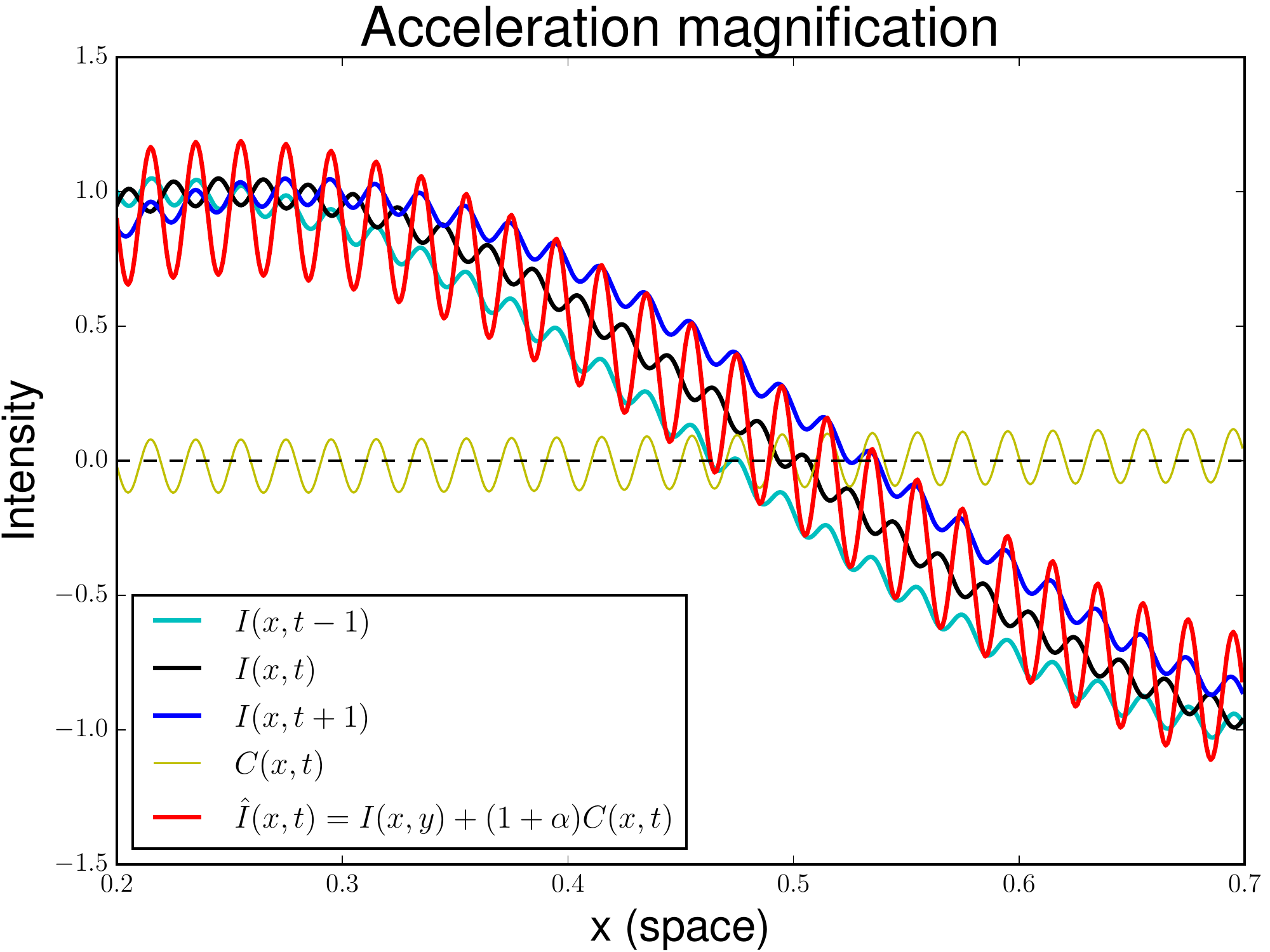} \\
 (a) & (b) 
 \end{tabular}	
\caption{\small 
Illustration of a 1 $D$ signal where small motions undergo a larger translation for linear magnification and acceleration magnification. The signal $I(x,t)$ is shown for 3 time instants, $\{t-1,t,t+1\}$. The red line shows the magnification results for a factor $\alpha=3$. (a) For first-order methods, 
the linear filter $B(x,t)$ is magnified and added to the original signal $I(x,t)$. Note that all motions are magnified, both small and large. (b) Acceleration magnification uses a temporal acceleration filter $C(x,t)$ which is magnified and added to the original signal $I(x,t)$. By assuming local linearity of the large translation motion, the translation has little effect on the magnification and only the small, non-linear, motions are magnified. This allows our method to magnify small changes of moving objects or scenes recorded with a moving camera.
}
\label{fig:1dOrder}
\end{figure*}

We take inspiration from prior work on linear Eulerian video magnification 
\cite{wuSIGGRAPH12eulerian,wadhwaSIGGRAPH13phase}. 
Linear magnification algorithms estimate and magnify subtle video changes --- pixel intensity or motion changes --- 
at fixed image locations, temporally. 

To illustratively compare our method to linear methods~\cite{wuSIGGRAPH12eulerian,wadhwaSIGGRAPH13phase} we consider a $1D$ signal with small motion changes 
under a larger translation motion, see \fig{1dOrder}.  

For input signal $I(x,t)$ at position $x$ and time $t$, the linear method assumes a displacement function $\delta(t)$ such that $I(x,t)= f(x + \delta(t))$. The goal is to synthesize $\hat I(x,t)= f(x + (1+\alpha) \delta(t))$ where $\alpha$ is the magnification factor. 

Assuming that the signal at time $t$ can be decomposed by a first-order Taylor series expansion around $x$ gives:
\begin{equation}
	I ( x , t ) \approx f(x) + \delta (t) \frac{\partial f(x)}{ \partial x},
	\label{eq:taylor}
\end{equation}
where the first-order term $\delta (t) \frac{\partial f(x)}{ \partial x}$ gives the linear change in signal over time. 

The linear magnification method uses a temporal bandpass filter 
$B(x,t)$ tuned to measure the desired video changes to be magnified:
\begin{equation}
	B (x,t) = \delta (t) \frac{\partial f(x)}{ \partial x}.
	\label{eq:filterOrder1}
\end{equation}
The magnified signal $\hat I (x,t)$ with a factor $\alpha$ is then:
\begin{equation}
	\hat I(x,t) = I(x,t) + \alpha B(x,t),
	\label{eq:velocity}
\end{equation}
which relates to the first-order term in the Taylor expansion:
\begin{equation}
	\hat I(x,t) \approx f(x) + (1 + \alpha) \delta (t) \frac{\partial f(x)}{ \partial x}.
\end{equation}
For details, see~\cite{wuSIGGRAPH12eulerian}.

Linear methods~\cite{wuSIGGRAPH12eulerian,wadhwaSIGGRAPH13phase} measure all motion changes: small motions and large motions. 
The bandpass filter $B(x,t)$ measures the magnitude of a change, and it does not discriminate if the change is big or small. Thus, all translational motion will be magnified. 
In~\fig{1dOrder}(a) we show the effect of large motions on linear magnification.  As the figure illustrates, linear methods are sensitive to large motions such as camera or object motion.

\subsection{Video Acceleration Magnification}
Rather than magnifying all temporal changes we magnify the deviation of change.
For example, if an object moves in one direction, then we enhance every small deviation from that direction. 
This includes the special case of an object that does not move, where deviations from no motion will be magnified. By assuming that the large object motion is approximately linear at the temporal scale of the small changes, we can disregard all linear motion. We do not magnify linear changes: we magnify accelerations.

For the 1 $D$ input signal $I(x,t) = f(x + \delta(t))$ at position $x$ and time $t$, we define its magnified counterpart as:
\begin{equation}
\hat I(x,t)= f(x + (1 + \beta)\delta(t)).
\end{equation}
Our goal is to model the magnified signal $\hat I(x,t)$ based on second-order changes.
Decomposing the magnified signal in a second order Taylor series around $x$ yields:
\begin{align}
	\hat I(x,t) & \approx  f(x) + (1 + \beta) \delta(t) \frac{\partial f(x)}{ \partial x} + \nonumber \\
		& + (1 + \beta)^2 \delta(t)^2 \frac{1}{2} \frac{\partial^2 f(x)}{ \partial x^2}.
	\label{eq:magni}
\end{align}
If we consider only the linear term of the magnified signal, we can define this as: 
\begin{align}
	\hat I(x,t)_{\text{linear}} & \approx  f(x) + (1 + \beta) \delta(t) \frac{\partial f(x)}{ \partial x} + 0.
\end{align}
However, here we aim at magnifying the non-linear part of the  signal, the acceleration. We can obtain this by subtracting the linear motion: 
\begin{align}
	\hat I(x,t)-\hat I(x,t)_{\text{linear}} & \approx (1 + \beta)^2 \delta(t)^2 \frac{1}{2} \frac{\partial^2 f(x)}{ \partial x^2},
\end{align}
where for simplicity we take $(1+\beta)^2 = \alpha$, with $\alpha > 0$.

Let $C(x,t)$ be the result of applying a temporal acceleration filter to $I(x,t)$ at every position $x$, then we capture the second-order offset:
\begin{equation}
	C(x,t)  = \delta(t)^2 \frac{1}{2} \frac{\partial^2 f(x)}{ \partial x^2},
\end{equation}
which we can multiply with $\alpha$ as the magnification factor 
\begin{equation}
	\hat I(x,t) = I(x,t) + \alpha C(x,t).
\end{equation}	
We focus on magnifying second-order signal changes: acceleration. 
In~\fig{1dOrder}(b) we show the effect of large motions on acceleration magnification.  
As the figure illustrates, our method only magnifies the small motion and is robust to large motions such as camera or object motion.

\subsection{Temporal Acceleration Filtering}

Acceleration is the second temporal derivative of the signal $I(x,t)$. 
To take a second-order derivative of the discrete video signal we use a Laplacian filter. 
The Laplacian is the second-order derivative of the Gaussian filter and it allows us to take an exact derivative of a smoothed discrete signal. The Gaussian is the only filter that does not introduce spurious resolution~\cite{koenderinkBioCyb4structOfIm}
and due to the linearity of the operators~\cite{koenderinkTPAMI92genericNeighOperators} the relation between the Laplacian and the second derivative of the signal is: 
\begin{equation}
\frac{\partial^2 I(x,t)}{\partial t^2} \otimes G_{\sigma}(t) = I(x,t) \otimes \frac{\partial^2 G_{\sigma}(t)}{\partial t^2},
\end{equation}
where $\otimes$ is convolution and $G_{\sigma}(t)$ is a Gaussian filter with variance $\sigma^2$ and $\frac{\partial^2 G_{\sigma}(t)}{\partial t^2}$ is the Laplacian.

The $\sigma$ parameter of the Gaussian allows for selecting the observation scale of the frequency to magnify~\cite{lindebergBook13scaleSpaceInCV,mikolajczyk2001indexing}.  For setting the observation scale, we denote the desired frequency by $w$ and we select a temporal window in the video that is equal to our target frequency as 
$\frac{r}{4w}$, 
where $r$ denotes the video frame rate. We center the temporal window on the current video frame. Subsequently, following \cite{mikolajczyk2001indexing}, 
we find the scale of the Laplacian kernel as: $\sigma = \frac{r}{4w\sqrt{2}}$. 
%

\subsection{Phase-based Acceleration Magnification}

For magnifying motion information, rather than intensity changes over time, we use as a starting point the successful work of~\cite{wadhwaSIGGRAPH13phase} where phase information is magnified by using the linear method of~\cite{wuSIGGRAPH12eulerian}. We use acceleration magnification in the phase domain to magnify non-linear motions. 


Motion can be represented by a phase shift. 
For a given input signal $f(x)$ with displacement $\delta(t)$ at time $t$, we can decompose the signal by Fourier series 
as sum of sinusoids over all frequencies $w$:
\begin{equation}
f(x+ \delta (t)) = \sum_{w=-\infty}^{\infty} A_w e^{iw(x+\delta(t))},
\end{equation}
where the global phase information at frequency $w$ for the displacements $\delta(t)$ is $\phi_w = w(x+\delta (t))$. 

Spatially localized phase information of an image over time is related to local motion~\cite{fleet1990computation} and is used for magnifying motions in the phase domain linearly~\cite{wadhwaSIGGRAPH13phase}. This motion magnification method uses the complex steerable pyramid~\cite{portilla2000parametric} to separate the image signal into multi frequency bands and orientations. The pyramid contains a set of filters $\Psi_{w,\theta}$ at various scales $w$, and orientations $\theta$. The local phase information of the 2D image $I(x,y)$ is given by: 
\begin{alignat}{2}
	(I(x,y) \otimes \Psi_{w,\theta})(x,y) &= A_{w,\theta}(x,y) e^{i\phi_{w,\theta}(x,y)},
\end{alignat}
where $\otimes$ is convolution, $A_{w,\theta}(x,y)$ is the amplitude and $\phi_{w,\theta}$ the corresponding phase at scale $w$ and orientation $\theta$. 


The phase information $\phi_{w,\theta}(x,y,t)$ at a given frequency $w$, and orientation $\theta$ and frame $t$, 
is magnified in our proposed approach by temporally filtering the phase $\phi_{w,\theta}(x,y,t)$ with a Laplacian:
\begin{alignat}{2}
	\hat{\phi}_{w,\theta}(x,y, t) &= \phi_{w,\theta}(x,y, t) + \alpha C_\sigma (\phi_{w,\theta}(x, y, t)), \\
	C_\sigma (\phi_{w,\theta}(x,y,t) ) &= \phi_{w,\theta}(x,y,t) \otimes \frac{\partial^2 G_\sigma(x,y,t)}{\partial t^2},
\end{alignat}
where $\otimes$ is convolution and $C_\sigma (\cdot)$ represents the temporal Laplacian filter with scale $\sigma$.

Due to the periodicity of the phase between $[-\pi, \pi]$, there is an interval ambiguity that may be present: a small increase to a value slightly less then $2\pi$ at time $t$ may cause the phase to become slightly bigger than $0$ at time $t+1$. This causes artifacts in the convolution with the Laplacian. We correct for this using phase unwrapping~\cite{kitahara2015algebraic}.



\section{Results}

\subsection{Experimental Setup}
\begin{table}
	\centering
	\label{experiment_video_table}
	\begin{tabular}{lllll}
		\toprule
		Video                 & $\alpha$ & $w$ (Hz) & Gaussian $\sigma$ & FPS \\ \midrule
		Light bulb   & 20     & 60       & 2.95      & 1000     \\
		Baby        & 100    & 2.5      & 6.63       & 30 \\ 
		Gun          & 8      & 20       & 4.24      & 480      \\ \midrule
		
		Synthetic ball & 8 & 2 & 5.30 & 60 \\
		
		Cat toy      & 4      & 3        & 1.41      & 240      \\
		Parkinson-1  & 3      & 3        & 2.12      & 30       \\
		Parkinson-2  & 4      & 3        & 2.12      & 30       \\
		Drone        & 5      & 5        & 1.06      & 30       \\
		Water bottle & 4      & 2        & 2.83      & 30       \\ \bottomrule
	\end{tabular}
	\caption{Parameters for all videos. 
		``Light bulb'' and ``Gun'' are from~\cite{wadhwaSIGGRAPH13phase}, the rest is new.}
	\label{tab:setup}
\end{table}
We evaluate our proposed method on real videos as well as synthetic ones with ground truth magnification.
We set the magnification factor $\alpha$, and the frequency of the change to be magnified as given in table~\ref{tab:setup}.
For all videos we process the video frames in YIQ color space.
We provide these videos as well as additional videos depicting our magnification method in the supplementary material.\\[5px]
\noindent \textbf{Motion Magnification.} 
We use the complex steerable pyramid \cite{portilla2000parametric} with half-octave bandwidth filters and eight orientations. 
We decompose each frame into magnitude and phase, and convolve with our proposed kernel over the phase signal temporally.\\[5px]
\noindent \textbf{Color Magnification.}
We decompose each video frame into multiple scales using a Gaussian pyramid, 
and we magnify the intensity changes only in the third level of the pyramid, similar to \cite{wuSIGGRAPH12eulerian}.

\subsection{Real-Life Videos}

\subsubsection{Comparison on Existing Videos}

\begin{figure} 
	\centering
	\includegraphics[width=0.45\textwidth]{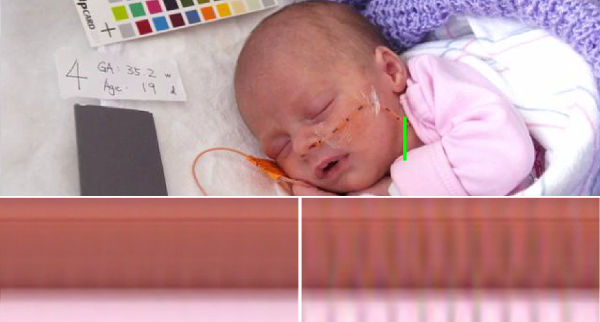}
	\caption{\small Intensity magnification on a static video. We indicate with a green stripe the locations at which we temporally sample the video. Note that our method is well able to magnify the intensity for videos without large motions.}
		\label{fig:baby}
\end{figure}

\begin{figure} 
	\centering
	\includegraphics[width=0.45\textwidth]{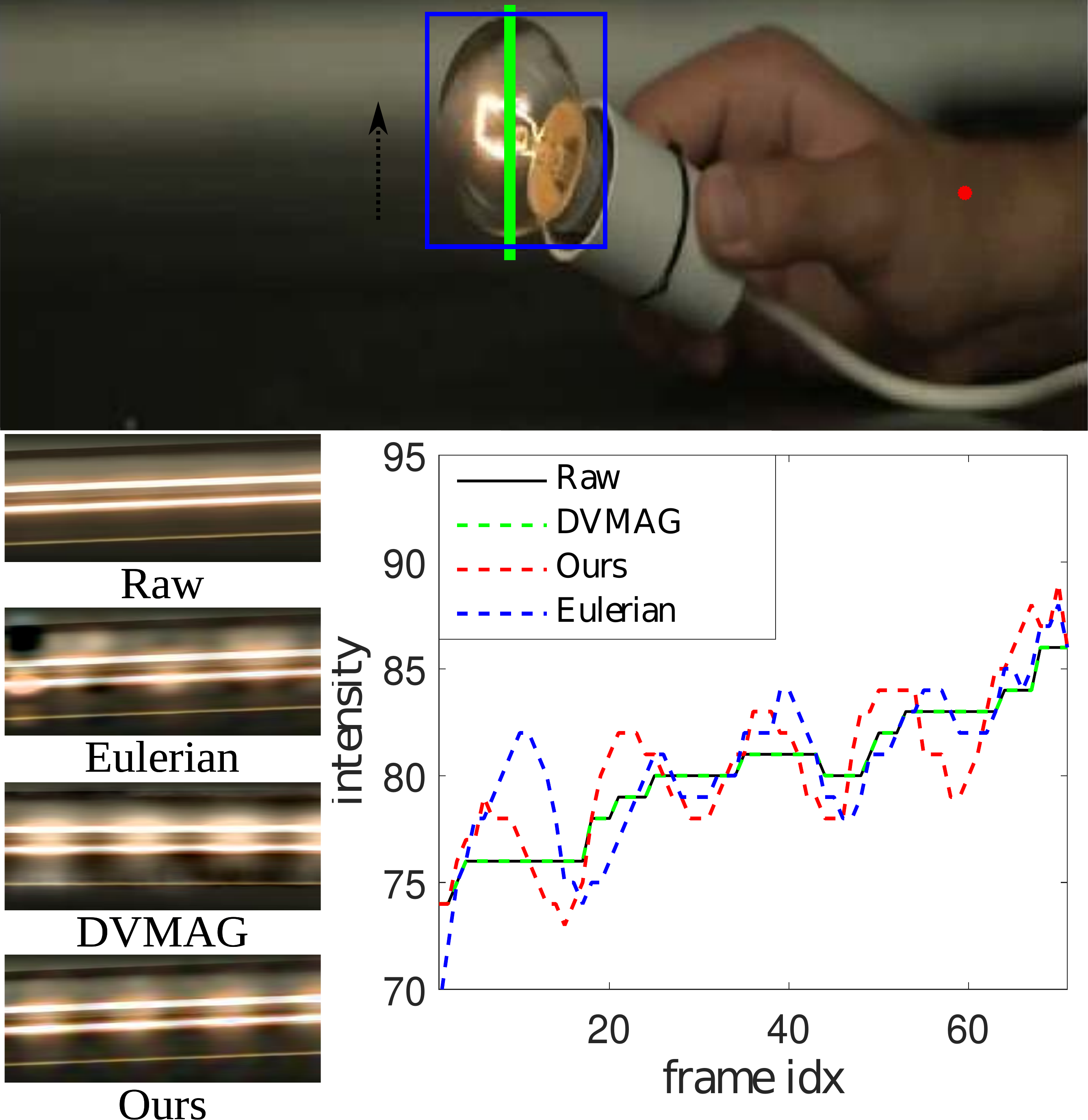}
	\caption{\small Intensity magnification. Note that the hand holding the light bulb moves upwards. 
	We indicate with a green stripe the locations at which we temporally sample the video.
	We show the original intensity change, 
	the Eulerian \cite{wuSIGGRAPH12eulerian} intensity magnification,
	the DVMAG \cite{elgharibCVPR15largeMotionMagnify}, where the blue region shows the user input area in which changes are magnified, 
	and our proposed acceleration magnification. 
	We also show the intensity changes over time in the hand area reflecting the light of the bulb. 
	The intensity changes are measured at the indicated red dot. 
	Our proposed method manages to magnify the intensity changes of the light bulb, but it also captures the intensity changes in the hand 
	cause by the reflection of the light. 
	}
	\label{fig:light}
\end{figure}

As a first experiment we show in \fig{baby} we show that our method can also magnify changes when there is no motion in the video.

Figure~\ref{fig:light} shows a person holding a light bulb while the hand moves upwards. 
The intensity variations in the light bulb are hardly visible. 
The Eulerian-based method \cite{wuSIGGRAPH12eulerian} reveals the intensity changes, but creates additional artifacts. 
DVMAG \cite{elgharibCVPR15largeMotionMagnify} relies on a user-input region around the bulb and therefore does not magnify the small reflections on the hand. 
Our proposed method not only magnifies the intensity variations of the light bulb without manual masking, 
but also magnifies the intensity changes of the hand, caused by the reflection of the light, as shown in the plot on the right of Figure~\ref{fig:light}.

\begin{figure*} 
	\centering
	\begin{tabular}{cccc}
		\includegraphics[width=0.22\textwidth]{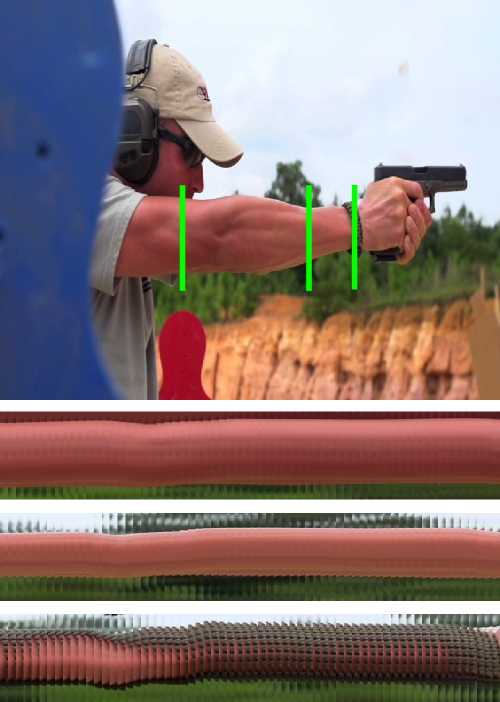} & 
		\includegraphics[width=0.22\textwidth]{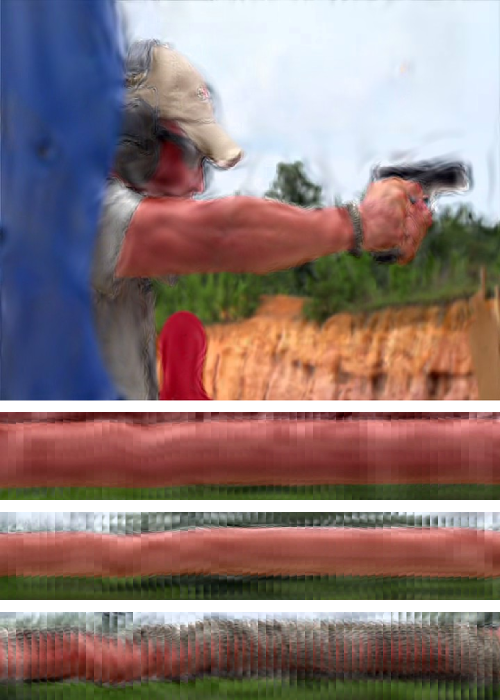} &
		\includegraphics[width=0.22\textwidth]{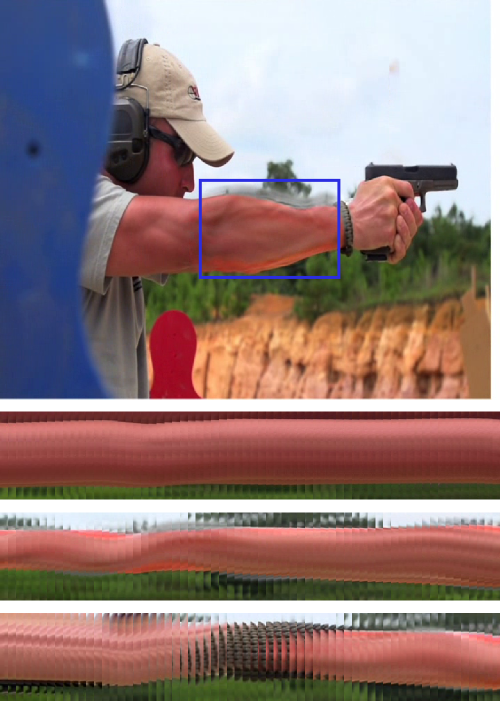} & 
		\includegraphics[width=0.22\textwidth]{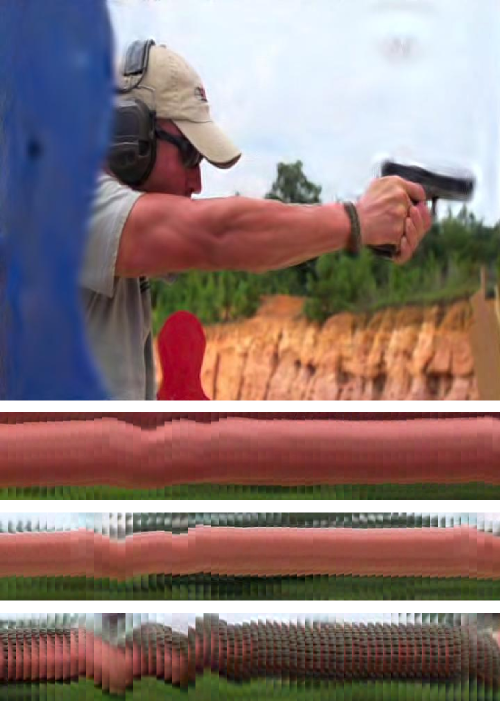} \\
		(a) Raw video. & (b) Phase-based \cite{wadhwaSIGGRAPH13phase}. & (c) DVMAG \cite{elgharibCVPR15largeMotionMagnify}. & (c) Ours.\\
	\end{tabular}
	\caption{\small Sports use-case: sports video analysis with motion magnification.
	(a) Original video frame. We indicate with three green stripes the locations at which we temporally sample the video.
	(b) Phase-based based motion magnification \cite{wadhwaSIGGRAPH13phase}. 
	(c) The DVMAG \cite{elgharibCVPR15largeMotionMagnify} results with user annotated areas indicated in blue. 
	(c) Our proposed acceleration magnification. 
	This figure shows a gun shooting sequence, where the recoil of the gun induces movement in the arm muscles. 
	DVMAG only magnifies the motion within the user annotated region, while the Eulerian based method results in large artifacts.
	Our proposed method magnifies he arm motion without inducing blurring and artifacts. 
	}
	\label{fig:gun_result}
\end{figure*}

Figure \ref{fig:gun_result} shows various motion magnification results for a gun shooting sequence. 
Due to the strong recoil, subtle motion in the arm muscles can be recovered. 
We record the motion of the forearm, upper limb, and the bracelet in the spatio-temporal slices indicated with three green lines over the original video. 
The phase-based motion magnification proposed in \cite{wadhwaSIGGRAPH13phase} induces large artifacts due to the strong arm movement. 
The DVMAG \cite{elgharibCVPR15largeMotionMagnify} relies on a user annotated region where the motion is magnified.
Therefore, the magnification performance depends on the user input, as seen in the figure.
Our method magnifies the muscle movement of the complete arm without creating artifacts and without the need for user input.

\begin{figure*} 
	\centering
	\begin{tabular}{cccc}
		\multicolumn{2}{c}{\includegraphics[width=0.45\textwidth]{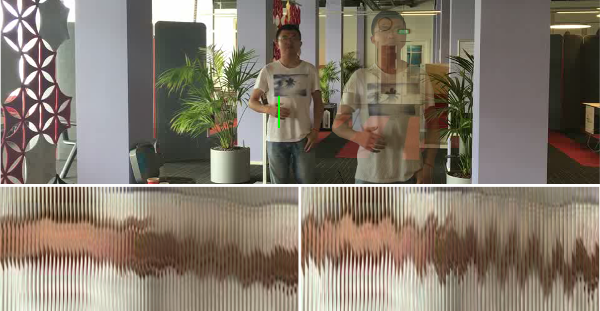}} & 
		\multicolumn{2}{c}{\includegraphics[width=0.45\textwidth]{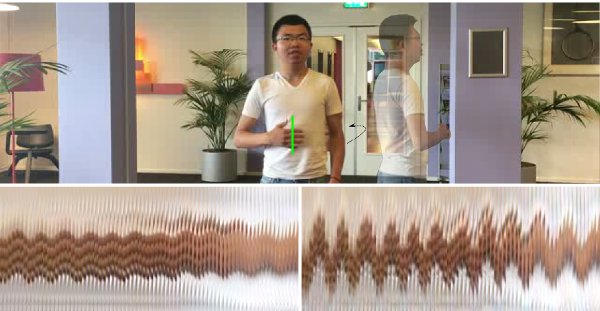}}\\ 
		\hspace{25px}(a) Raw video. & (b) Our magnification & \hspace{25px}(a) Raw video. & (b) Our magnification.\\
	\end{tabular}
	\caption{\small Medical use-case: hand tremor magnification. The left example (Parkinson-1) has the person walking towards the screen. 
	The right example (Parkinson-2) has the person do a $3D$ rotation. We overlay 2 frames of the video to visualize how the person moves.  
	(a) Original video frames. We indicate with a green stripe the locations at which we temporally sample the video.
	(b) Our proposed acceleration magnification. 
	We manage to amplify the motion in the arm of the person while the person is moving towards the camera and even under a 3D rotatation. This is possible because the scale of the body motion is considerably larger than the scale of the hand tremor. 
	}
	\label{fig:parkinson}
\end{figure*}
\subsubsection{Additional Videos with Large Object Motion}
Figure 1 shows a toy moving on the table while vibrating with a high frequency.
The goal of the experiment is to magnify the vibration while not creating artifacts and blurring. 
Our proposed method manages to achieve this by magnifying the motion at the pixels that have a non-zero acceleration, thus amplifying the 
vibration of the toy and ignoring the motion along the trajectory of the toy on the table.

In figure~\ref{fig:parkinson} we consider a medical use case in which a person walks towards screen --- zooming, and a video in which a person is rotating in $3D$, while having a tremor motion present in the right arm. 
Our proposed approach is able to magnify the tremor of the arm without introducing considerable artifacts and blurring in the rest of the areas. 
\begin{figure} 
	\centering
	\begin{tabular}{cc}
		\multicolumn{2}{c}{\includegraphics[width=0.45\textwidth]{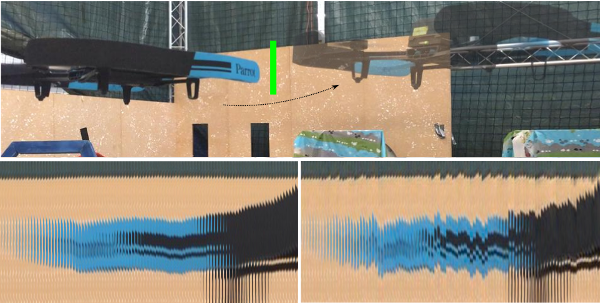}}\\ 
		\hspace{25px}(a) Raw video. & (b) Our magnification.\\
	\end{tabular}
	\caption{\small Mechanical use-case: analyzing possible mechanical failures from motion. A drone oscillating while flying in a cluttered environment.   
	(a) Original video frames. We indicate with a green stripe the locations at which we temporally sample the video.
	(b) Our proposed acceleration magnification. 
	Our proposed magnification method is able to amplify the oscillations of the drone without being affected by the background clutter.
	}
	\label{fig:drone}
\end{figure}

\begin{figure} 
	\centering
	\begin{tabular}{cc}
		\multicolumn{2}{c}{\includegraphics[width=0.45\textwidth]{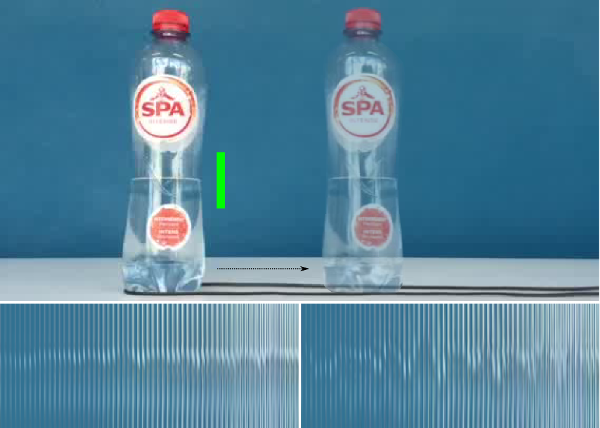}}\\ 
		\hspace{25px}(a) Raw video. & (b) Our magnification.\\
	\end{tabular}
	\caption{\small The water fluctuating in a bottle while the bottle is being pulled sideways on a smooth surface. 
	(a) Original video frames. We indicate with a green stripe the locations at which we temporally sample the video.
	(b) Our proposed acceleration magnification. 
	Our propose magnification method is able to amplify the fluctuations in the water level while not adding substantial blur.
		}
	\label{fig:water}
\end{figure}

In figure~\ref{fig:drone} we show our results on a mechanical stability quality control application where a drone is oscillating while flying in a cluttered environment. 
Moreover, in \fig{water} we show a transparent bottle with water being pulled on a smooth table --- the level of water in the bottle fluctuates. 
Our method is able to correctly magnify the desired motion --- oscillation of the drone and fluctuations of the water level, despite the challenging setup of 
background clutter and transparent elements whose motion must be magnified. 

	
\subsection{Controlled Experiments}
\begin{figure}
	\centering 
	\includegraphics[width=0.45\textwidth]{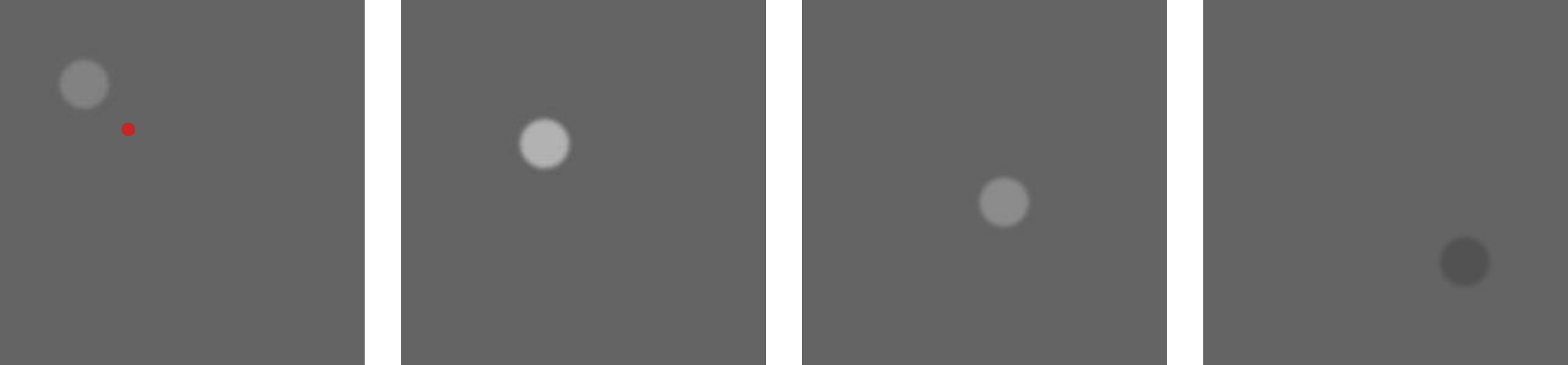} 
	\caption{ \small 
	Synthetic Video. 
	A ball with intensity varying while moving from top-left corner to the bottom-right. 
}
	\label{fig:experiment_ball}
\end{figure}

\begin{figure}  
	\hspace{-5px}
	\includegraphics[width=0.50\textwidth]{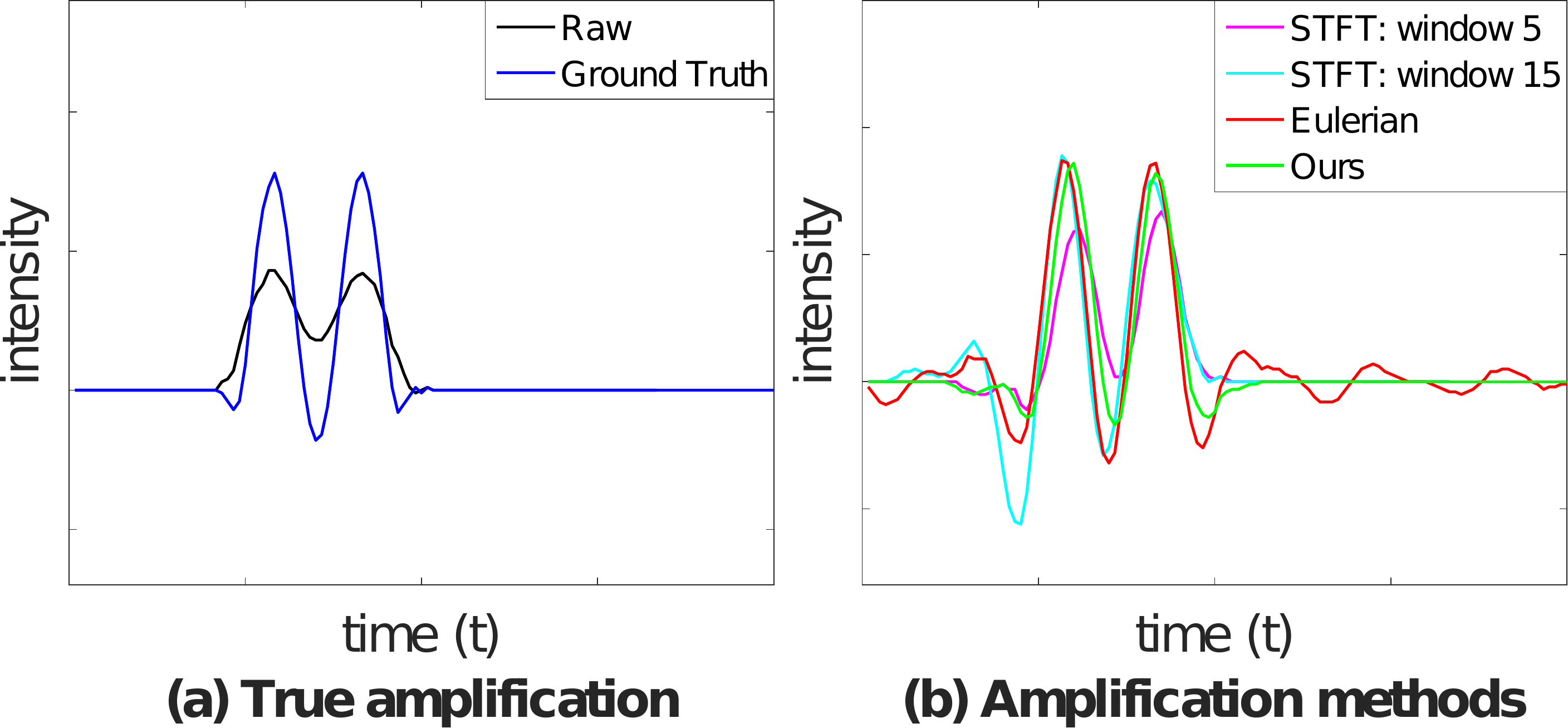} 
	\caption{ \small 
		(a) We record the change in intensity temporally at the value of the red point indicated in the left frame of figure \ref{fig:experiment_ball}. 
		The black curve shows the original intensity values, while the blue curve shows the ground truth magnification. 
		(b) Signal magnification result for our method, 
		the Eulerian method \cite{wuSIGGRAPH12eulerian}, and STFT (Short Term Fourier Transform) with window sizes 5 and 15.
		Our method generated a signal magnification closer to the ground truth magnification, while not creating 
		additional artifacts.
	}
	\label{fig:ball_intensity_plot}
\end{figure}

In figure~\ref{fig:experiment_ball} we show a synthetic ball which moves diagonally on the screen from the top-left corner to bottom-right corner, 
with its intensity fluctuating in certain frequency. 
We set the radius of ball as 10 pixels. 
The ball moves with 1 pixel/frame. 
We model the intensity changes as a sine wave, with a maximum intensity change of 20.
The intensity frequency is $2$ cycle/sec, and we set the frame rate to $60$ frame/sec. 
For ground truth magnification, we amplify the intensity changes 4 times without changing any other parameters. 
For all methods, we first apply a Gaussian pyramid and only magnify the third pyramid level with amplification factor $8$. 

Figure \ref{fig:ball_intensity_plot} shows magnification results for a set of considered baselines. 
We compare with an ideal filter of $1.5-2.5$ Hz from the Eulerian magnification method in \cite{wuSIGGRAPH12eulerian} which uses the whole video. To make this a more fair baseline we also use this method with the STFT (Short Term Fourier Transform)  with a temporal window of frame sizes 5 and 15.
The Eulerian approach generates background artifacts due to the bandpass filter which uses the complete temporal length of the video. 
STFT partially alleviates this problem, artifacts being removed outside the temporal window.
However, it generates larger artifacts inside the temporal window. 
For a smaller window size the intensity changes are magnified less, because at a coarse frequency resolution in Fourier domain more signals are filtered out.
Our method generates an intensity magnification that closely resembles the ground truth, without introducing artifacts.

\begin{figure}
	\hspace{-15px}
	\begin{tabular}{c@{\hspace{3px}}c}
		\includegraphics[width=0.250\textwidth]{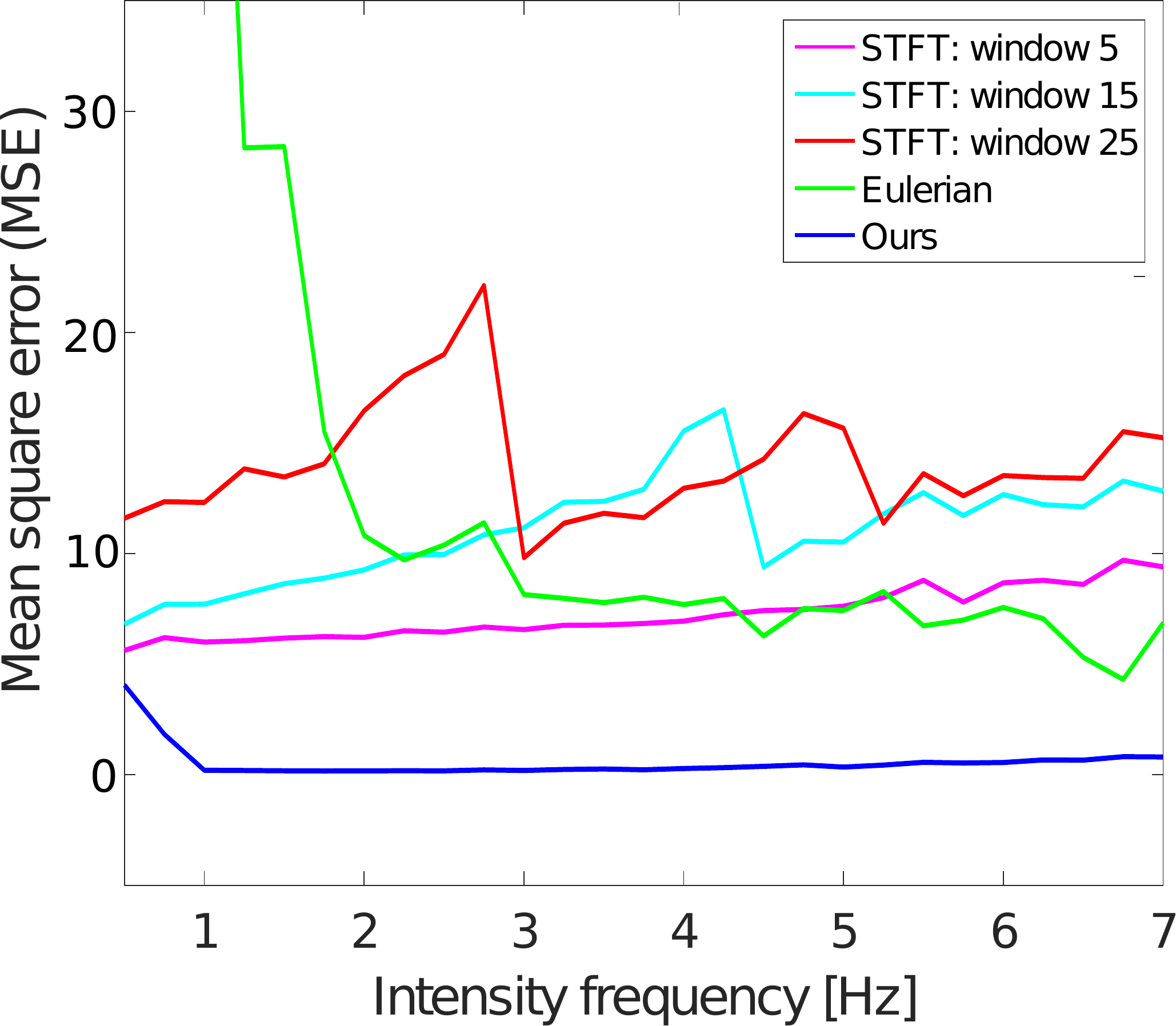} &
		\includegraphics[width=0.250\textwidth]{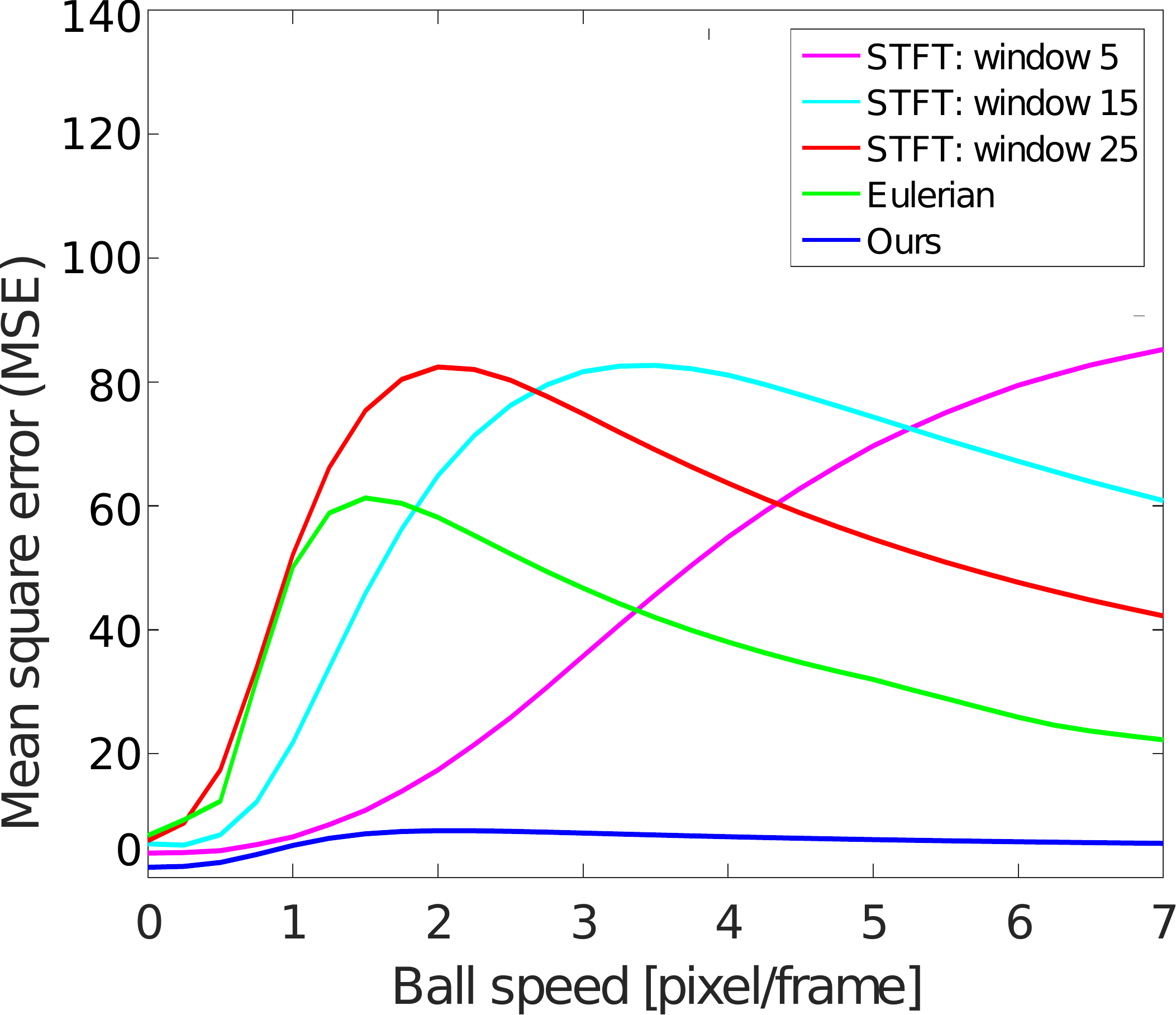}\\
		(a) & (b)\\
	\end{tabular}
	\caption{ \small (a) Error while increasing intensity frequency. (b) Error while increasing object speed. }
	\label{fig:MSE}
\end{figure}

We analyze the effect of the intensity frequency on the magnification methods. The ball speed is fixed to $0.5$ pixel/frame, and we vary the intensity frequency from $0.5$ Hz to $7$ Hz in increments of $0.25$ Hz while keeping other parameters unchanged. We estimate MSE (Mean Square Error) between the predicted intensity and the ground truth intensity magnification, measured over the whole image in all frames.
Results are given in figure~\ref{fig:MSE}.(a). The error of the Eulerian method \cite{wuSIGGRAPH12eulerian} decreases with the increase in intensity frequency. 
This is because the ideal bandpass filter in the frequency domain is able to measure more periods of the signal at high frequencies. The STFT methods, perform well when the corresponding temporal window contains precisely one cycle of the intensity change. 
For example, for an STFT with window size 25, there is a drop in MSE around the frequency $2.5$ Hz, 
while for STFT with window size 15, the drop is at $4$ Hz. Our method is sensitive to low frequencies, where the signal barely fits in the temporal window. For higher frequencies the method stabilizes and outperforms the others.

For analyzing the effect of the speed on the magnification methods we fix the intensity frequency at $2$ Hz, and increase the ball speed with increments of $0.25$ from $0$ to $7$ pixel/frame while keeping other parameters unchanged.  
In figure~\ref{fig:MSE}.(b) it shows that the Eulerian approach \cite{wuSIGGRAPH12eulerian} and the STFT methods have trouble for speeds around 1.5 pixel/frame. 
For most methods, MSE decreases slowly with the increase in ball speed. The high error for the lower frequencies is mostly due to blurring effects outside the ball. When increasing the speed of the ball, less intensity changes are available to measure. Our proposed method has a similar behavior, albeit at a better performance level then others. 

\section{Discussion}
\noindent \textbf{Limitations of our approach.} 
A disadvantage of a second-order filter is the need for two zero-crossings for estimating it. 
Thus, the phase measurement has to be made for at least 3 frames inside the spatial aperture at hand. 
The spatial aperture corresponds to various levels in the multi-resolution complex spatial pyramid, where the highest resolution has the smallest aperture. 
Thus, for fast moving objects, fine texture will blur the easiest, exemplified by motion boundary artifacts. 
Other limitations of our method are the assumption that the large motion should be linear. 
Therefore, if nonlinear motion is present in the background, this will also be magnified causing blur. 
The same effect occurs when multiple motions with different speeds are present.
However, assuming a single linear large motion, our method performs well when 
the object is moving --- slightly rotating or zooming (\fig{parkinson}), and when the camera is moving --- provided linear, non-shaky, camera motion.   
These results can be found in the supplemental.

\noindent \textbf{Performance improvement over state-of-the-art.}
Our results are similar to \cite{elgharibCVPR15largeMotionMagnify} on the manually drawn masked regions of DVMAG \cite{elgharibCVPR15largeMotionMagnify}. 
For non-masked regions, the results differ. 
For a moving camera we do not change the video resolution, nor introduce white boundaries as \cite{elgharibCVPR15largeMotionMagnify}. 
For \fig{gun_result}, our method incorrectly blurs the gun, whereas \cite{elgharibCVPR15largeMotionMagnify} does not. 
Yet, we magnify the bracelet and shoulder, revealing a pattern that \cite{elgharibCVPR15largeMotionMagnify} misses. 
On \fig{light} we magnify the hand color, which \cite{elgharibCVPR15largeMotionMagnify} does not. 
We sidestep the problem of drawing a mask for unseen motion that is not yet visible. 
Results in the supplementary material on the ``parking gate" and ``eye" videos of \cite{elgharibCVPR15largeMotionMagnify} look similar for ``eye", 
while for ``parking gate" the multiple non-linear motions in the background cause blurring effects. 
We expect the difference to be further emphasized when testing \cite{elgharibCVPR15largeMotionMagnify} on our videos, 
as non-planar 3D rotation (\fig{cattoy} and \fig{parkinson}) is a serious issue for optical-flow homography alignment as used in \cite{elgharibCVPR15largeMotionMagnify}, 
however the implementation of \cite{elgharibCVPR15largeMotionMagnify} is not available.

Our results, original videos and code implementation can be found at:
\href{https://acceleration-magnification.github.io}{https://acceleration-magnification.github.io}.

\section{Conclusions}

We present a method for magnifying small changes in the presence of large motions. Standard video magnification algorithms~\cite{wuSIGGRAPH12eulerian,wadhwaSIGGRAPH13phase} cannot handle large motion while the  concurrent DVMAG method~\cite{elgharibCVPR15largeMotionMagnify} requires user annotations, optical flow, and temporal alignment. We are not bounded by such constraints and can magnify unconstrained videos.  

We magnify acceleration by measuring deviations from linear motion. We do this by linking the response of a second-order Gaussian derivative to spatial acceleration. 

We demonstrate our approach on synthetic and several real-world videos where we do better, and/or require less user intervention than other methods. Our real-world videos show the potential of our method in the medical domain (Parkinson-I and Parkinson-II), 
in sports (Gun), and in mechanical stability quality control (Drone).

{\small
	\bibliographystyle{ieee}
	\bibliography{zhangCVPR2017}
}

\end{document}